\documentclass[letterpaper, 10 pt, conference]{ieeeconf}  % Comment this line out if you need a4paper

\usepackage{xcolor}
\newif\ifpaperfinal
\paperfinalfalse
\ifpaperfinal
\else

\usepackage{adjustbox}
\usepackage{booktabs}
\usepackage{subcaption}
\usepackage[hyphens]{url}
\graphicspath{ {./images/} }
\usepackage{siunitx}
\usepackage{hyperref}
\usepackage{algorithm}
\usepackage[noend]{algpseudocode}
\usepackage{amsmath}
\usepackage[numbers]{natbib}
\usepackage{subcaption}
\usepackage{graphicx}

\title{\LARGE \bf
Advanced Autonomy on a Low-Cost Educational Drone Platform 
}

\IEEEoverridecommandlockouts

\author{Luke Eller$^{1*}$, Th\'eo Gu\'erin$^{1*}$, Baichuan Huang$^{1*}$, Garrett Warren$^{1*}$, Sophie Yang$^{1*}$, \\Josh Roy$^{1}$, Stefanie Tellex$^{1}$% <-this % stops a space
\thanks{$^{*}$\textbf{Denotes equal contribution}}%
\thanks{$^{1}$Department of Computer Science, Brown University, Providence, RI, 02912, USA. Correspondence:  {\tt\small  \{luke\_eller, theo\_guerin, baichuan\_huang, garrett\_warren, sophie\_yang, josh\_roy, stefanie\_tellex\}@brown.edu}}%
}

\begin{document}

\maketitle
\thispagestyle{empty}
\pagestyle{empty}

%%%%%%%%%%%%%%%%%%%%%%%%%%%%%%%%%%%%%%%%%%%%%%%%%%%%%%%%%%%%%%%%%%%%%%%%%%%%%%%%
\begin{abstract}
  PiDrone is a quadrotor platform created to accompany an introductory robotics course. Students build an autonomous flying robot from scratch and learn to program it through assignments and projects. Existing educational robots do not have significant autonomous capabilities, such as high-level planning and mapping. We present a hardware and software framework for an autonomous aerial robot, in which all software for autonomy can run onboard the drone, implemented in Python. We present an  Unscented Kalman Filter (UKF) for accurate state estimation. Next, we present an implementation of Monte Carlo (MC) Localization and FastSLAM for Simultaneous Localization and Mapping (SLAM). The performance of UKF, localization, and SLAM is tested and compared to ground truth, provided by a motion-capture system. Our evaluation demonstrates that our autonomous educational framework runs quickly and accurately on a Raspberry Pi in Python, making it ideal for use in educational settings.
\end{abstract}

%%%%%%%%%%%%%%%%%%%%%%%%%%%%%%%%%%%%%%%%%%%%%%%%%%%%%%%%%%%%%%%%%%%%%%%%%%%%%%%%
\section{INTRODUCTION}

Substantial increase in demand in the field of robotics demonstrates the need for autonomous educational platforms. The International Data Corporation predicts that global spending on robotic technologies---and drones in particular---will grow annually over the next several years at a compound rate of nearly 20 percent, which is a massive opportunity for continued innovation in the field \cite{idc}. However, the plethora of knowledge and technical skills required for this growing domain is a considerable barrier to entry. This paper focuses on making advanced autonomy accessible to individuals with no robotics experience. We build on the low-cost educational platform introduced in \cite{brand2018pidrone} by adding advanced algorithms for state estimation, localization, and SLAM. The algorithms are implemented in Python and documented in novel course projects.

Improvement in drone technology has made many commercial autonomous aircraft widely available. These include the Tello EDU~\cite{dji} from DJI which provides high-level APIs for education and the Skydio R1~\cite{skydio} which provides an SDK for developing. These commercial drones primarily target high-level programming aspects and are not open-source. There are some open-source drone platforms for advanced college-level courses; however, these are not suitable for students with less background in engineering and robotics \cite{brand2018pidrone}. The PiDrone platform and course were created to fill this gap. 

\begin{figure}
  \includegraphics[width=0.8\linewidth]{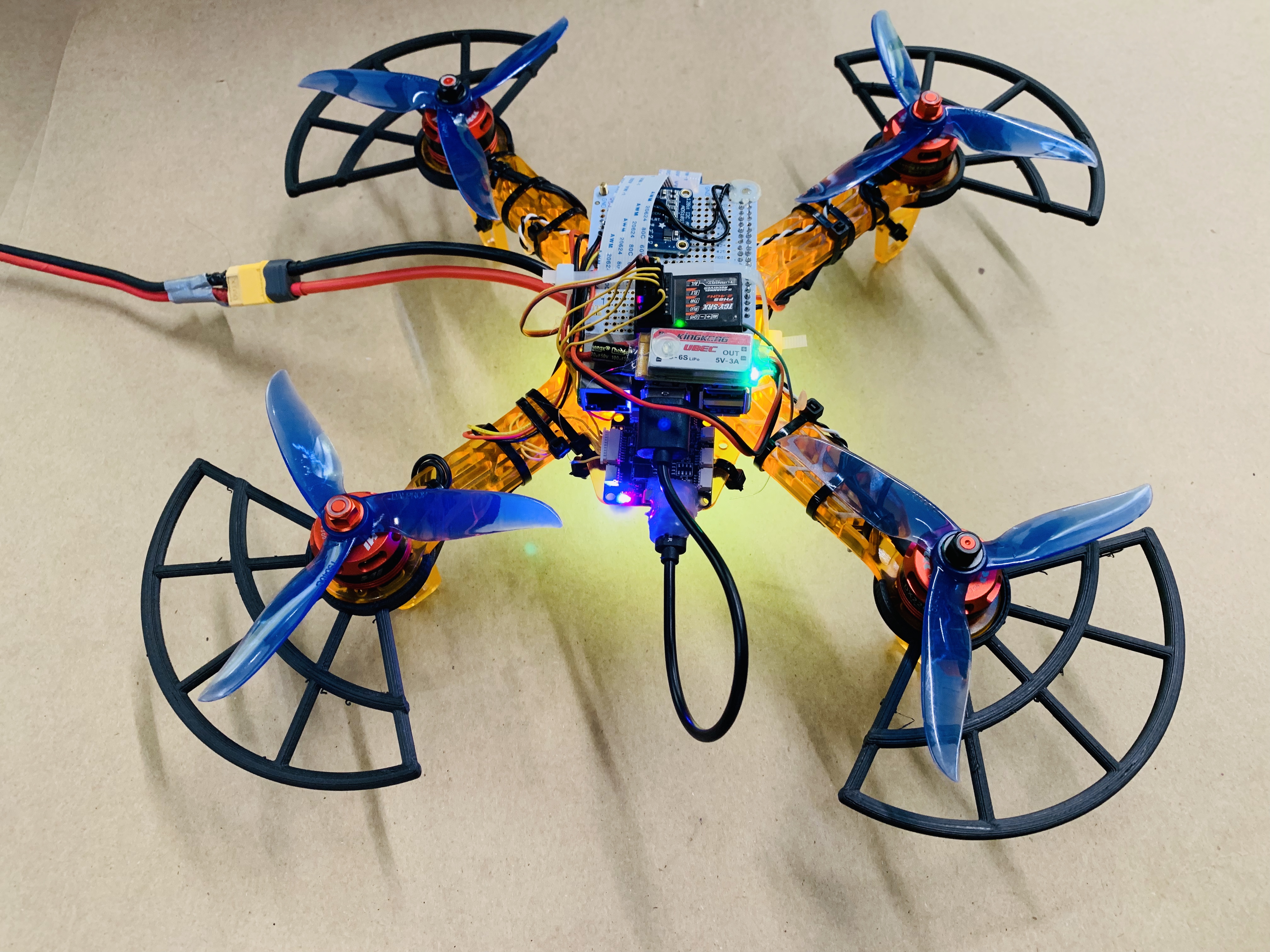}
  \centering
  \caption{PiDrone Hardware Platform}
  \label{fig:drone}
  \vspace{-4mm}
\end{figure}

For complex tasks such as mapping, precise position control, or trajectory following it is necessary to have precise state estimates of velocity and position. In this paper, we present an onboard UKF to better estimate state, as well as localization and SLAM implementations to generate more accurate position estimates. Mathematical descriptions of these algorithms provide a formal description of the state, observation, and control models used to obtain good performance. Corresponding course material includes these algorithms and was successfully taught to undergraduate students at Brown University in 2018 \cite{website} \cite{textbook} \cite{build_manual}.

In Section \ref{section:performance}, we quantitatively evaluate the performance of the UKF and localization running entirely onboard the drone's Raspberry Pi, and FastSLAM running offboard on a separate base-station computer with ROS installed. Even though FastSLAM cannot run onboard in an efficient way, students can still learn and implement the algorithm and then delegate the computation to another machine to increase performance. The accuracy of the position estimates obtained by the UKF, localization, and SLAM compared to ground-truth measured by a motion-capture system exemplifies the drone's ability to serve as an educational platform for these algorithms. The fact that the algorithms can all be written in Python increases the accessibility of these algorithms, which is especially crucial for an introductory robotics course.

%%%%%%%%%%%%%%%%%%%%%%%%%%%%%%%%%%%%%%%%%%%%%%%%%%%%%%%%%%%%%%%%%%%%%%%%%%%%%%%%

\section{ARCHITECTURE}

\subsection{Hardware}

The PiDrone follows a similar parts list to \cite{brand2018pidrone} with a few improvements, making it safer and easier to build while remaining under \$225. Notable hardware changes include the use of a Raspberry Pi HAT, an add-on board, to improve the accessibility of the build process and the robustness of the built drone \cite{pihat}. Soldering directly to the pins of the Raspberry Pi is a difficult task for introductory students. Instead, the Raspberry Pi HAT allows students to solder to pads rather than pins. We also add lightweight, open-source, 3D printed propeller guards to the platform as seen in Fig. \ref{fig:drone}. These guards provide additional safety for students and researchers.

\subsection{Software}
\begin{figure}
  \includegraphics[width=\linewidth,height= 4cm]{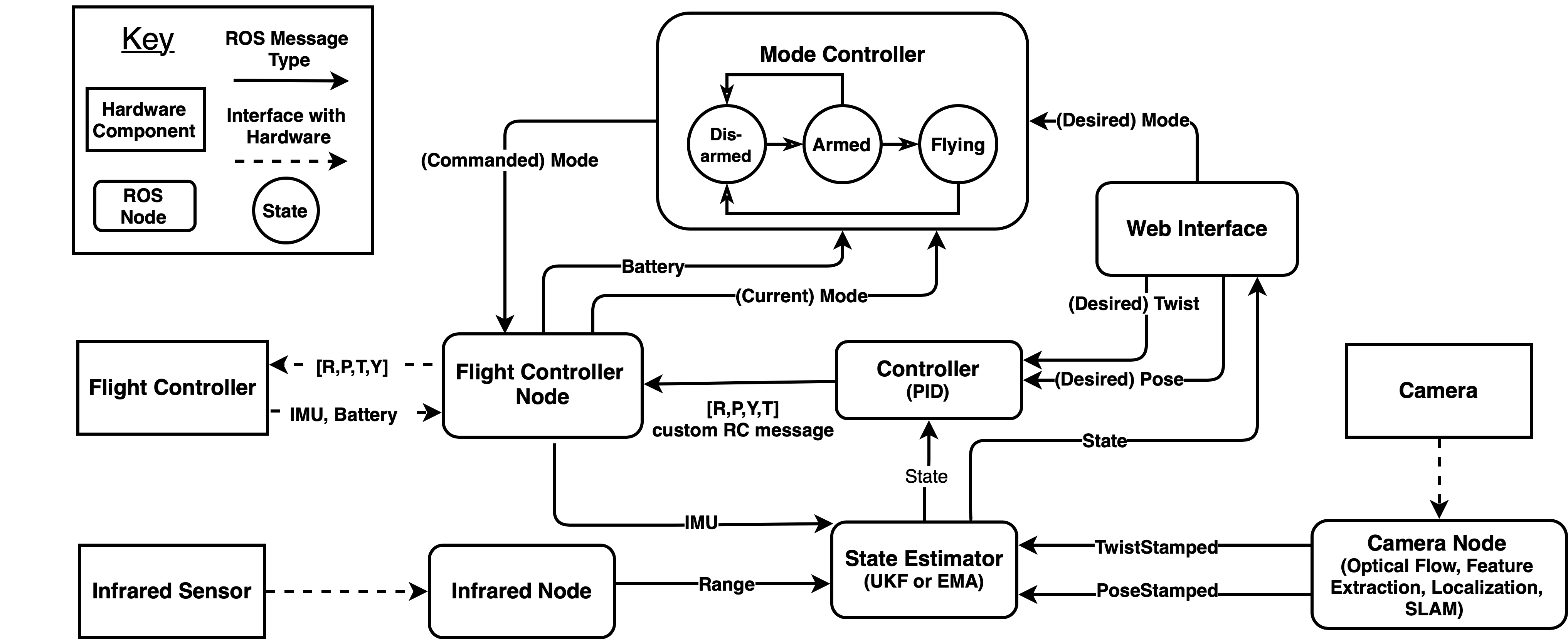}
  \centering
  \caption{Software Architecture Diagram}
  \label{fig:softwarearchitecture}
  \vspace{-4mm}
\end{figure}

To maximize ease of use of the PiDrone as an educational and research platform, the software architecture is organized as shown in Fig. \ref{fig:softwarearchitecture}. The modular architecture provides a plug-and-play software environment for the drone. The ease of swapping core software components including control algorithms, state estimators, sensor interfaces, and user inputs is demonstrated concretely through the corresponding educational course. Students implement and substitute in their own scripts for PID control, SLAM, and UKF state estimation without compromising the stability of the rest of the system. This modularity enables researchers to easily implement and evaluate a new state estimation or control algorithm or add a new sensor to the drone.

%%%%%%%%%%%%%%%%%%%%%%%%%%%%%%%%%%%%%%%%%%%%%%%%%%%%%%%%%%%%%%%%%%%%%%%%%%%%%%%%

\section{STATE ESTIMATION}

\subsection{Unscented Kalman Filter}
The UKF algorithm and the related Extended Kalman Filter (EKF) algorithm are the industry standards for state estimation of nonlinear systems, used by open-source projects such as Cleanflight and Betaflight, as well as commercial products such as the Crazyflie~\cite{cleanflight}~\cite{crazyflie}~\cite{betaflight}. The PiDrone course covers both the mathematics and use cases of the Kalman Filter, EKF, and UKF, using \textit{Probabilistic Robotics} as reference material~\cite{Thrun2005}; however, student work is centered on the UKF. By implementing the UKF themselves, students taking this course will be prepared to understand and work with state estimation systems either in academia or in industry.

The EKF is widely used for Micro Aerial Vehicles to estimate state, including acceleration, velocity, and position~\cite{flyingcars}. It combines data from multiple sensors to form state estimates, fitting the use case for the PiDrone. The UKF achieves better estimation performance while remaining no more computationally intensive than the EKF~\cite{ukf}.

The PiDrone software stack implements the UKF due to its performance benefits. Additionally, unlike the EKF, the UKF does not require computing derivatives, making the algorithm more accessible to introductory robotics students.

As the mathematical details of the UKF are quite substantial for an introductory robotics course, we do not require students to implement many of the computations and instead make use of the Python library FilterPy, which has an accompanying online textbook that is presented to students if they desire deeper understanding of the UKF \cite{labbe_kalman}. The project in which students implement the UKF focuses more on the higher-level design decisions and specifications that are necessary to adapt the general UKF algorithm to a particular robotic system such as the PiDrone. Some of these UKF design specifications for our drone, such as which state variables to track and the state transition and measurement functions, are presented.

The Unscented Kalman Filter implemented on the drone calculates prior state estimates, computed in the \textit{prediction} step, and posterior state estimates, computed in the \textit{measurement update} step. Due to the computational complexity of the UKF, two variants were developed to run on the drone: first, a UKF consisting of a simple one-dimensional model of the drone's motion to estimate its position and velocity along the vertical axis, and second, a model that encompasses three spatial dimensions of motion. To differentiate between these two UKFs, we refer to them by the dimension of their state vectors: the simpler model tracks a two-dimensional (2D) state vector, while the more complex model estimates a seven-dimensional (7D) state vector.

\subsubsection{Two-Dimensional UKF}

The 2D UKF has the state vector $\mathbf{x}_t$ shown in Equation (\ref{eq:2dukf}), which tracks position and velocity along the $z$-axis.
\begin{equation} \label{eq:2dukf}
\mathbf{x}_t=\begin{bmatrix}
z &
\dot z
\end{bmatrix}^\mathsf{T}
\end{equation}

To carry out the prediction step of the UKF algorithm, we use a control input $\mathbf{u}_t=\begin{bmatrix}\ddot z\end{bmatrix}$, which is the linear acceleration along the $z$-axis measured by the Inertial Measurement Unit (IMU) onboard the drone. It has been shown that in certain cases the incorporation of IMU data in the measurement update step may be more accurate \cite{imu_predict_vs_update}; however, for relative ease of implementation, we have chosen to treat accelerations as control inputs. This choice is an example of a design decision that students are made aware of when implementing the UKF.

The state transition function, with inputs of the previous state estimate $\mathbf{x}_{t-\Delta t}$, the control input $\mathbf{u}_t$, and the time step $\Delta t$, is shown in Equation (\ref{eq:2dukf_state_transition}) following one-dimensional kinematics \cite{tellexquadrotors}.
\begin{equation} \label{eq:2dukf_state_transition}
g(\mathbf{x}_{t-\Delta t}, \mathbf{u}_t, \Delta t) = \begin{bmatrix}
1 & \Delta t \\
0 & 1
\end{bmatrix}\mathbf{x}_{t-\Delta t} + \begin{bmatrix}
\frac{1}{2}(\Delta t)^2 \\
\Delta t
\end{bmatrix}\mathbf{u}_t
\end{equation}

After a prior state estimate is calculated with the state transition function, the algorithm moves to the measurement function to incorporate sensor measurements. For the 2D UKF, we only consider the drone's downward-facing infrared range sensor---which provides a range reading $r$---in the measurement update step, so our measurement vector is given by $\mathbf{z}_t = \begin{bmatrix}r\end{bmatrix}$. The measurement function, then, transforms the state vector into measurement space by selecting the $z$ position component, as shown in Equation (\ref{eq:2dukf_measurement}) \cite{tellexquadrotors} .
\begin{equation} \label{eq:2dukf_measurement}
h(\mathbf{\bar x}_t) = \begin{bmatrix}
1 & 0
\end{bmatrix}\mathbf{\bar x}_t
\end{equation}

To gain an understanding of the role of the covariance matrices involved in the UKF algorithm, students collect data to characterize the sample variance $\sigma_r^2$ of their infrared range sensor. The software stack includes simple data simulation to aid students and researchers as they tune parameters such as the covariance matrices.

\subsubsection{Seven-Dimensional UKF}

The 7D UKF tracks motion in three spatial dimensions with the state vector in Equation (\ref{eq:7dukf}), as well as the drone's yaw angle $\psi$.
\begin{equation} \label{eq:7dukf}
\mathbf{x}_t = \begin{bmatrix}
x &
y &
z &
\dot x &
\dot y &
\dot z &
\psi \end{bmatrix}^\mathsf{T}
\end{equation}

We define a control input $\mathbf{u}_t = \begin{bmatrix}\ddot x^b & \ddot y^b & \ddot z^b\end{bmatrix}^\mathsf{T}$ populated by linear accelerations from the IMU in the drone's body frame. These accelerations are transformed into the global coordinate frame by taking into account the drone's estimated yaw from the state vector, as well as its roll and pitch angles, which are filtered by the IMU. This transformation is carried out with quaternion-vector multiplication as shown in Equation (\ref{eq:quat_vec}), where $\mathbf{q}$ is the quaternion that rotates a vector from the body to the global frame. The use of quaternions---which appear frequently in robotics---in the state transition function offers students an introduction to this mathematical representation.
\begin{equation} \label{eq:quat_vec}
\mathbf{u}^g_t = \mathbf{q} \cdot \mathbf{u}_t \cdot \mathbf{q}^*
\end{equation}

The resulting global-frame accelerations $\mathbf{u}_{\ddot x^g}, \mathbf{u}_{\ddot y^g}, \mathbf{u}_{\ddot z^g}$ of $\mathbf{u}^g_t$ are used in the state transition function in Equation (\ref{eq:7dukf_state_transition}) \cite{tellexquadrotors}.

\begin{equation} \label{eq:7dukf_state_transition}
\begin{gathered}
g(\mathbf{x}_{t-\Delta t}, \mathbf{u}_t, \Delta t) = \mathbf{x}_{t-\Delta t} + \begin{bmatrix}
\dot x \Delta t + \frac{1}{2}\mathbf{u}_{t,\ddot x^g}\left( \Delta t \right)^2 \\
\dot y \Delta t + \frac{1}{2}\mathbf{u}_{t,\ddot y^g}\left( \Delta t \right)^2 \\
\dot z \Delta t + \frac{1}{2}\mathbf{u}_{t,\ddot z^g}\left( \Delta t \right)^2 \\
\mathbf{u}_{t,\ddot x^g} \Delta t \\
\mathbf{u}_{t,\ddot y^g} \Delta t \\
\mathbf{u}_{t,\ddot z^g} \Delta t \\
0
\end{bmatrix}, 
% \text{where } \dot x, \dot y, \dot z \text{ are components of } \mathbf{x}_{t-\Delta t}
\end{gathered}
\end{equation}
where $\dot x$, $\dot y$, $\dot z$ are components of $\mathbf{x}_{t-\Delta t}$.

The measurement update step uses the measurement vector $\mathbf{z}_t = \begin{bmatrix}r & x & y & \dot x & \dot y & \psi_{\text{camera}}\end{bmatrix}^\mathsf{T}$, where $r$ is the infrared slant range reading, $x$ and $y$ are planar position estimates from the downward-facing camera, $\psi_{\text{camera}}$ is the camera's yaw estimate, and $\dot x$ and $\dot y$ are velocity estimates from the camera's optical flow. Equation (\ref{eq:7dukf_measurement}) shows the measurement function, which uses roll $\phi$ and pitch $\theta$ angles to transform altitude into slant range.
\begin{equation} \label{eq:7dukf_measurement}
h(\mathbf{\bar x}_t) = \begin{bmatrix}
\frac{\mathbf{\bar x}_{t,z}}{\cos \theta \cos \phi} &
\mathbf{\bar x}_{t,x} &
\mathbf{\bar x}_{t,y} &
\mathbf{\bar x}_{t,\dot x} &
\mathbf{\bar x}_{t,\dot y} &
\mathbf{\bar x}_{t,\psi}
\end{bmatrix}^\mathsf{T}
\end{equation}

If the planar position estimates originate from localization or SLAM, which already act as filters on raw camera data, then it may not be notably beneficial to apply a UKF on top of these estimates; however, the capability for researchers to incorporate such measurements exists in the 7D UKF implementation, and the higher-dimensional state space offers educational insights to students. For the above reason as well as the computational overhead brought on by estimating seven state variables, at present, the 2D UKF is preferred when attempting stable flight. We analyze the performance of the 2D UKF in the Robot Performance section.

\subsection{Localization}

The UKF provides state estimation via sensor fusion, but most mobile robots---the PiDrone included---do not include a sensor to directly measure position~\cite{Thrun2005}~\cite{brand2018pidrone}. 
As such, the PiDrone software stack implements both localization and SLAM using the particle filter and FastSLAM algorithms described in \textit{Probabilistic Robotics}~\cite{Thrun2005} and the FastSLAM paper~\cite{FastSLAM}. 

The implementation of Monte Carlo Localization is based on the particle filter according to~\citet{Thrun2005} with a customized  \texttt{sample\_motion\_model}, \texttt{measurement\_model}, and particle update process to meet our specific hardware setup. We took the idea of \texttt{keyframe} from~\citet{leutenegger2013keyframe} to only perform \texttt{measurement\_model} updates when necessary: namely, when the drone has moved a significant distance. We use ORB feature detection from OpenCV \cite{opencv_library} to provide the sensor data. We implement a simple version of the localization algorithm that is easy to understand for students and fast enough to run onboard the Raspberry Pi.

Our implementation of \texttt{sample\_motion\_model} is a simplified odometry motion model as shown in Equations (\ref{eq:motion1}) and (\ref{eq:motion2}). The $\delta_x, \delta_y, \delta_{\theta}$ are the translations and rotation, and the $\varepsilon_{\sigma_x^2}, \varepsilon_{\sigma_y^2}, \varepsilon_{\sigma_{yaw}^2}$ are the noise we added which are Gaussian zero-mean error variables with variances $\sigma_x^2, \sigma_y^2, \sigma_{yaw}^2$. The algorithm keeps track of particle motion, adding noisy translations and rotations to each particle in every frame. We tried using a velocity-based motion model where the velocity is provided by optical flow from the camera~\cite{brand2018pidrone}; however, the low-cost camera is influenced by many factors such as light and its reflection. Instead, we use the transform between two frames as parameters for the motion model. With feature detection and matching, we can compute the translation between two frames. Also, we assume the height is known which is provided by the infrared sensor. Therefore, we implemented a 2D localization algorithm.

\begin{equation}\label{eq:motion1}
    \begin{bmatrix}
    \hat{\delta}_{x} \\
    \hat{\delta}_{y} \\
    \hat{\delta}_{\theta}
    \end{bmatrix}
=
\begin{bmatrix}
    \delta_{x} \\
    \delta_{y} \\
    \delta_{\theta}
\end{bmatrix}
+
\begin{bmatrix}
    \varepsilon_{\sigma_x^2} \\
    \varepsilon_{\sigma_y^2} \\
    \varepsilon_{\sigma_{\theta}^2}
\end{bmatrix}
\end{equation}

\begin{equation}\label{eq:motion2}
    \begin{bmatrix}
    x' \\
    y' \\
    \theta'
\end{bmatrix}
=
\begin{bmatrix}
    x \\
    y \\
    \theta
\end{bmatrix}
+
\begin{bmatrix}
    \hat{\delta}_{x} \cdot \cos(\theta) - \hat{\delta}_{y} \cdot \sin(\theta) \\
    \hat{\delta}_{x} \cdot \sin(\theta) + \hat{\delta}_{y} \cdot \cos(\theta) \\
    \hat{\delta}_{\theta}
\end{bmatrix}
\end{equation}

The implementation of \texttt{measurement\_model} is feature-based with known correspondence as described in Chapter 6 of \textit{Probabilistic Robotics}~\citep{Thrun2005}. The algorithm computes the global position $(x', y', \theta')$ of particles based on $features$ from the current frame as shown in Equation (\ref{eq:measurement1}). The $features$ variable is the collection of features that are extracted from the current frame. To narrow down the feature matching space, the \texttt{compute\_location} function takes features from the current frame to compare with features that are close to the position of the particle. 

In the \texttt{sample\_motion\_model}, we add observation noise to the estimated position before we compute the likelihood of estimated position and the current position of the particle. However, the $\sigma^2$ in Equation (\ref{eq:measurement2}) is the variance for measurement which is different than Equation (\ref{eq:motion1}) which is for motion. These are empirically tuned values.

Also, we compute the likelihood $q$ of the estimated position of each particle in Equation (\ref{eq:measurement3}), simply trusting that most groups of features are unique (the drone flies over non-repeating textured surfaces~\cite{brand2018pidrone}), instead of using the likelihood of landmarks as described in~\cite{Thrun2005}. This process simplified the measurement model so that it can provide position estimation onboard while flying. 
\begin{equation}\label{eq:measurement1}
    x', y', \theta' = \texttt{compute\_location}(features)
\end{equation}
\begin{equation}\label{eq:measurement2}
    \begin{bmatrix}
    \hat{x}' \\
    \hat{y}' \\
    \hat{\theta}'
\end{bmatrix}
=
\begin{bmatrix}
    x' \\
    y' \\
    \theta'
\end{bmatrix}
+
\begin{bmatrix}
    \varepsilon_{\sigma_x^2} \\
    \varepsilon_{\sigma_y^2} \\
    \varepsilon_{\sigma_{\theta}^2}
\end{bmatrix}
\end{equation}
\begin{align}\label{eq:measurement3}
    q = & \texttt{prob}(\hat{x'}-x, \sigma_x)\notag \\
    & \cdot \texttt{prob}(\hat{y'}-y, \sigma_y)\notag \\
    & \cdot \texttt{prob}(\hat{\theta'}-\theta, \sigma_{\theta})
\end{align}

The update process of the Monte Carlo Localization is similar to that from \citet{Thrun2005}, except that we use a keyframe scheme: the \texttt{measurement\_model} is called only if the \texttt{sample\_motion\_model} has been processed a certain number of times or the estimated position of the drone is far from the estimate at the last time we performed the \texttt{measurement\_model} process. This allows the localization algorithm to run in a single thread onboard, which would be realizable for students with less multi-threading background. 

\subsection{Simultaneous Localization and Mapping} \label{sssec:num1}

FastSLAM, described in \citet{FastSLAM}, is a particle filter algorithm for Simultaneous Localization and Mapping. The dominant approach to this problem has been to use an EKF to estimate landmark positions. The chief problem with this approach is computational complexity: the covariance matrix of an EKF-SLAM with $N$ landmarks has at least $N^2$ entries. Instead, FastSLAM factors the posterior distribution by landmarks, representing each landmark pose with a single EKF. There are some more modern SLAM algorithms such as ORB-SLAM2 \cite{murORB2}, which may provide more accurate state estimation in 3D space; however, such algorithms are more complicated to understand for students. FastSLAM is simple to understand and a natural fit for the PiDrone platform as it directly builds off of Monte Carlo Localization. Particles in MC Localization are augmented by adding a landmark estimator (EKF) for each observed landmark to each particle. The motion model, keyframe scheme, and resampling methods are left unchanged from localization. The measurement model is replaced with a \texttt{map\_update} method to associate observed features with existing landmarks and assign a probability to particles, which each represent an estimate of the robot pose and map.

SLAM seeks to estimate the posterior distribution $p(\Theta, x^t | z^t, u^t)$ where $\Theta$ is the map consisting of N landmark poses $\Theta=\theta_1,...\theta_N$, $x^t$ is the path of the robot $x^t= x_1,...,x_t$, $z^t$ is the sequence of measurements $z^t= z_1,...,z_t$, and $u^t$ is the sequence of controls, $u^t= u_1,...,u_t$. The main mathematical insight of FastSLAM is the ability to factor this distribution by landmark as Equation (\ref{eq:slameq}).
\begin{equation}\label{eq:slameq}
    p(\Theta, x^t | z^t, u^t)=p(x^t | z^t, u^t)\Pi_n{p(\theta_n |x^t, z^t, u^t)}
\end{equation}
This approach is sound since individual landmark estimations are conditionally independent, assuming knowledge of the robot's path and correspondences between observed features and landmarks in the map. 

The factored posterior is realized with a 2D EKF to estimate each landmark pose in each particle of the filter. Newly observed ORB features \cite{opencv_library} are entered as landmarks into the map, and the EKFs of existing landmarks are updated when re-observed. The main difference of FastSLAM from MC Localization, then, is the \texttt{map\_update} step, described in Algorithm \ref{alg:map_update}, which associates newly observed features with existing landmarks and assigns a weight to each particle. 
Note that the $landmarks$ list consists of only landmarks within a small radius $r$ of the particle's pose, computed by line \ref{op2}. The method $get\_perceptual\_range()$ uses the drone's height to compute the radius of the largest circle within full view of the drone's camera. 
This ensures that the map represented by each particle is conditioned on the unique robot path represented by that particle. 
Also note that the landmark counter scheme ensures existing landmarks which are not matched to but lie within the camera's field of view are removed in line \ref{op6}. We determine match quality using Lowe's Ratio Test from \cite{LoweRatio} in line \ref{op1}. Line \ref{op3} finds the two best-matching landmarks for an observed feature using OpenCV's \texttt{knnMatch} \cite{opencv_library}. Finally, the weight of each particle is incremented for each landmark within its perceptual range. The weight is increased by an importance factor proportional to the quality of the match for re-observed landmarks in line \ref{op4}, and the weight is decreased for new landmarks in line \ref{op5}. The $threshold$ is defined as some constant between 0 and 1.

\begin{algorithm}
\caption{SLAM - Map Update}\label{alg:map_update}
\begin{algorithmic}[1]
\Procedure{map\_update}{$observed\_features$}

\State $r \gets get\_perceptual\_range()$ \label{op2}
\For {$p \in particle\_set$}
\State $landmarks \gets $ empty list

\For {$lm \in p.landmarks$}
\If{$dist(lm, p.pose) \leq r$}
\State $landmarks \gets landmarks \cup lm$
\EndIf
\EndFor

\For{$f\in observed\_features$}
\State $match1,match2 \gets \newline
    \hspace*{6em}best\_2\_matches(f,landmarks)$ \label{op3}
\If{$match1.dist > 0.7 \cdot match2.dist$} \label{op1}
\State $lm \gets initEKF(p.pose, f.pose)$
\State $lm.counter \gets 0$
\State $p.weight \gets \newline
    \hspace*{8em}p.weight + \log{(threshold)}$ \label{op5}
\Else 
\State $old\_lm \gets landmarks[match1\_idx])$
\State $lm \gets updateEKF( \newline
    \hspace*{8em} p.pose, f.pose, old\_lm)$
\State $importance \propto dist1 - dist2$
\State $p.weight \gets p.weight + importance$ \label{op4}
\EndIf
\State $p.landmarks \gets p.landmarks \cup lm$
\State $lm.matched \gets true$
\EndFor

\For{$lm \in p.landmarks$}
\If{$lm.matched$}
\State $lm.counter \gets lm.counter + 1$
\Else
\State $lm.counter \gets lm.counter - 1$
\EndIf
\If{$lm.counter < 0$}
\State $p.landmarks \gets p.landmarks \setminus lm$ \label{op6}
\EndIf
\EndFor

\EndFor
\EndProcedure
\end{algorithmic}
\end{algorithm}

Given the high density of features required for the motion model, PiDrone SLAM creates landmark-dense maps, resulting in slow map updates. It was found that multithreaded map updates are required to perform SLAM in real time (online SLAM) or else the robot path will get lost if the drone moves during map updates. Performing thread-safe updates to SLAM particles is challenging for introductory robotics students. Rather, we present a method for performing SLAM sequentially (offline SLAM), allowing students to implement the simple FastSLAM algorithm. We collect flight data on a flying drone, then perform SLAM offline on the saved data to build a map. The drone then flies, performing MC localization over the map created from SLAM; this runs in real time onboard the drone. The method shows that maps converge correctly by localizing over a map created by FastSLAM. \newline

%%%%%%%%%%%%%%%%%%%%%%%%%%%%%%%%%%%%%%%%%%%%%%%%%%%%%%%%%%%%%%%%%%%%%%%%%%%%%%%%

\section{ROBOT PERFORMANCE}\label{section:performance}
  
\begin{figure}
  \includegraphics[width=0.85\linewidth]{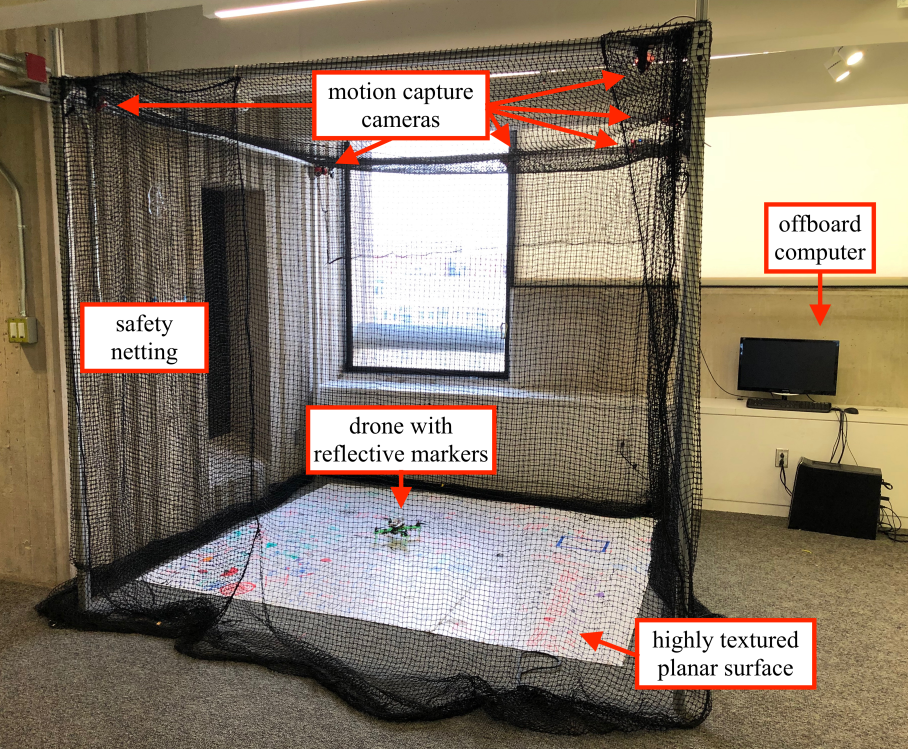}
  \centering
  \caption{Environmental setup used for evaluation.}
  \label{fig:flyingSpace}
  \vspace{0mm}
\end{figure}

The aim of our validation was to assess the performance of the
system relative to ground truth, in order to show that we are
accurately able to perform state estimation. Fig. \ref{fig:flyingSpace} depicts the testing environment which includes: the flying space enclosed in safety netting, the motion capture cameras, a highly textured planar surface to fly over, the drone with reflective markers, and the computer used for offboard computations.

\subsection{Unscented Kalman Filter}

\begin{figure}
\centering

\begin{subfigure}{\columnwidth}
\includegraphics[width=\columnwidth]{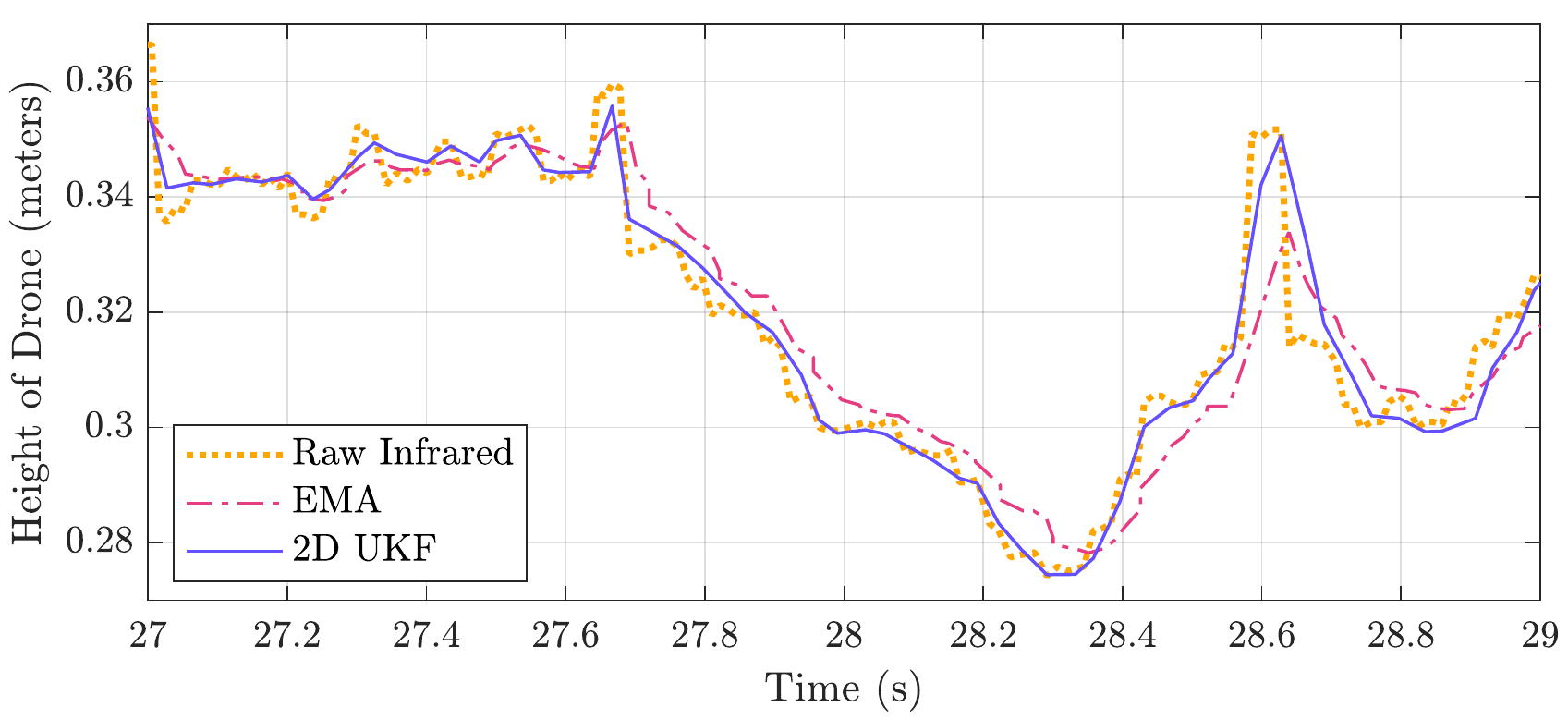}
\end{subfigure}

\begin{subfigure}{\columnwidth}
\includegraphics[width=\columnwidth]{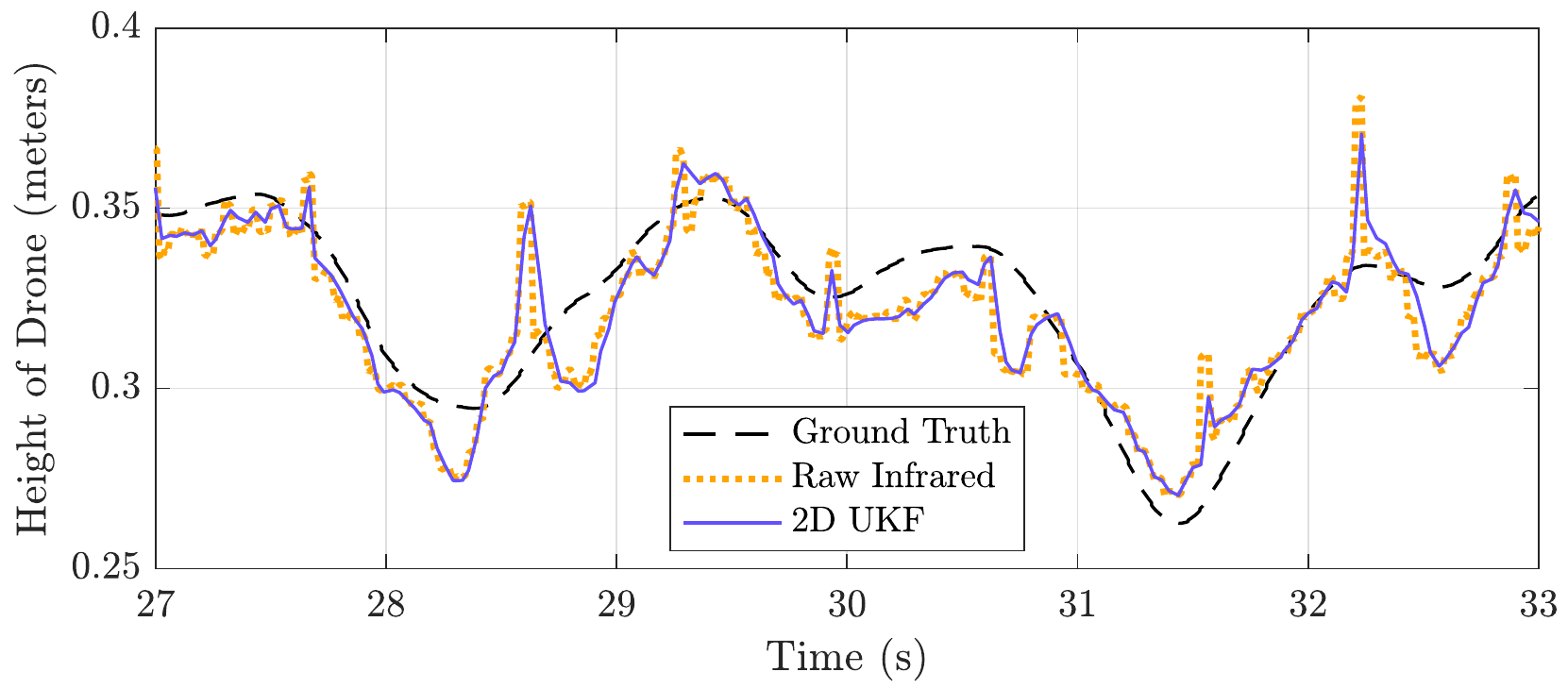}
\end{subfigure}

\caption{Height estimates of 2D UKF against other sources.}
\label{fig:2dukf_data}
\end{figure}

The 2D UKF was run onboard the Raspberry Pi with a predict-update loop executing at 30 Hz to provide the drone with estimates of its altitude. We compared these filtered estimates to the raw infrared range readings and to the smoothed readings from the Exponential Moving Average (EMA) filter, which was the filtering method used before the development of the UKF. However, the latency inherently introduced by an EMA filter was considered undesirable for a quadrotor. Fig. \ref{fig:2dukf_data} displays the results of a flight test in which the drone was commanded to hover in place. The 2D UKF curve follows the raw infrared readings with less latency than the EMA in quick ascents and descents. Additionally, the UKF estimates demonstrate smaller fluctuations than the noisy infrared sensor. Although the UKF estimates require more computation than a naive EMA, its benefits are apparent in Fig. \ref{fig:2dukf_data} and could be particularly useful for agile maneuvers.

The 2D UKF was also tested against ground-truth height provided by a motion capture system. The results of this comparison are shown in Fig. \ref{fig:2dukf_data}.

\begin{figure*}[!ht]
\centering
  \begin{subfigure}{5.5cm}
    \centering\includegraphics[width=5.5cm]{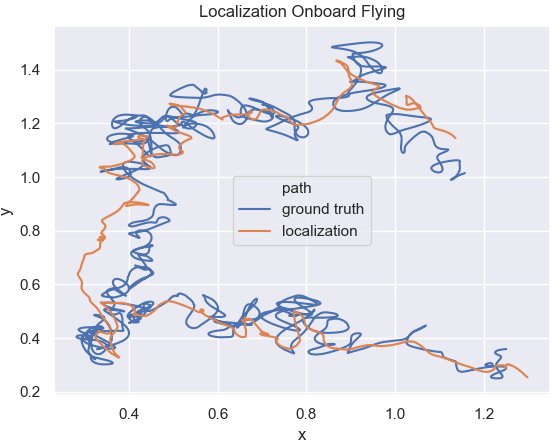}
    \caption{}\label{fig:locala}
  \end{subfigure}
  \begin{subfigure}{5.5cm}
    \centering\includegraphics[width=5.5cm]{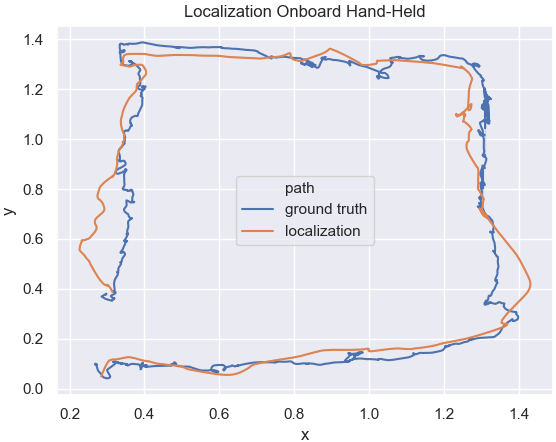}
    \caption{}\label{fig:localb}
  \end{subfigure}
  \begin{subfigure}{5.5cm}
    \centering\includegraphics[width=5.5cm]{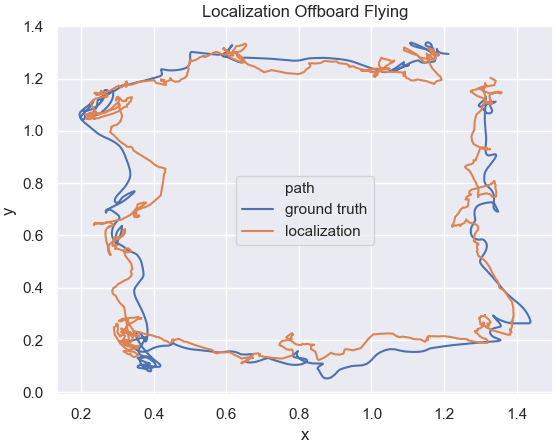}
    \caption{}\label{fig:localc}
  \end{subfigure}
  \caption{Localization algorithm runs in three conditions}
  \label{fig:local}
\end{figure*}

\subsection{Localization}
We set up the motion capture system to collect the $x$ and $y$ coordinates of the drone at 120 Hz. The onboard localization is running at 5 Hz with 40 particles, collecting 180 features at each frame along with 19,200 map features. The offboard localization is running at 12 Hz with 50 particles, collecting 250 features at each frame along with 48,000 map features. We use data from the motion capture system as the ground truth to validate our localization implementation. The drone was flying over a 1.67m $\times$ 1.65m textured surface, and we collected data while the drone was flying in a square. To prepare the map for localization, we took photos of the surface around 55 centimeters, and stitched them using Image Composite Editor from Microsoft \cite{image-comp-edit}. There are some distortions of the map image which causes offsets for a certain area, but our localization still works despite these distortions.

\begin{table}[t]
  \centering
    \begin{tabular}{rcccc}
        \toprule
          & Mean & Std & Maximum & Minimum\\
        \midrule
        Onboard Flying & 0.115 & 0.062 & 0.301 & 0.007\\
        Onboard Hand-Held & 0.096 & 0.049 & 0.261 & 0.010\\
        Offboard Flying & 0.074 & 0.042 & 0.190 & 0.007\\
        \bottomrule
    \end{tabular}
    \caption{Error between our Localization implementation and ground truth from motion capture system (meters).}
    \label{table:localizationacc}
    \vspace{0mm}
\end{table}

The localization running onboard while the drone is flying is compared to ground truth in Fig. \ref{fig:locala}. When running localization onboard the drone and flying, the Raspberry Pi spends less of its computational resources on the PID controller, resulting in less stable flight. This is due primarily to the computational load of the \texttt{measurement\_model}. To accurately evaluate the quality of the localization algorithm, we also collected data while moving the drone by hand. 
The results are shown in Fig. \ref{fig:localb}. We moved the drone slowly and steadily, and the localization algorithm estimated the position close to the ground truth. 
Those two images show that our simplified localization implementation is functional. Also, if students want to do further research with the PiDrone, the offboard version of localization provides higher accuracy and more stable flight as shown in Fig. \ref{fig:localc}, allowing other research code to run simultaneously. 

Table \ref{table:localizationacc} shows statistics for data we collected. We compared coordinates from our localization implementation to the ground truth and we pair them based on the Robot Operating System (ROS) timestamp. The error for one pair would be calculated as
$$
\text{error} = \lvert x - x' \rvert + \lvert y - y' \rvert,
$$
where $x, y$ are the planar coordinates from the localization algorithm and $x', y'$ are the planar coordinates from the motion capture system. Considering that the drone is flying and we did not account for the angle while the frame is captured, the mean error is acceptable. With more computing resources and stabler flying, the accuracy is higher.

\subsection{Simultaneous Localization and Mapping}

\begin{figure}
  \includegraphics[width=0.85\linewidth]{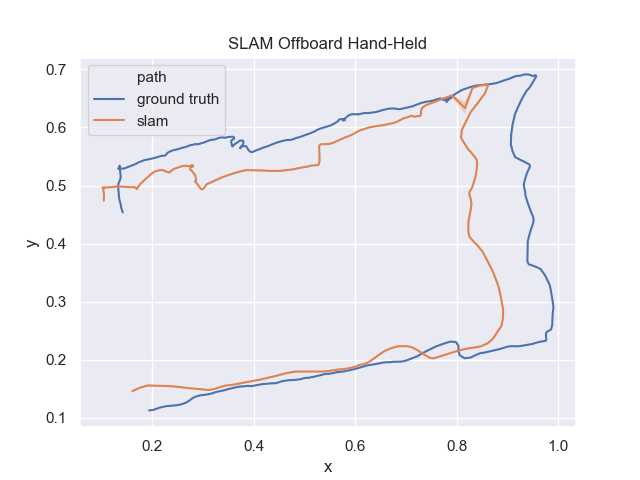}
  \centering
  \caption{Hand-Held Offboard Localization over SLAM-created map.}
  \label{fig:offboardSLAM}
  \vspace{-4mm}
\end{figure}

Data are collected from offline SLAM (see Section \ref{sssec:num1} for motivation) and compared to ground truth. Initially, the drone was moved by hand over a 1.67m x 1.65m highly textured planar surface, extracting 200 ORB features per frame at a rate of 30 Hz and saving them to a text file, as well as saving an infrared height reading with every image frame. Then, SLAM was performed with 40 particles onboard the grounded drone using the saved flight data, resulting in a map containing 5,108 landmarks. Finally, the drone was moved by hand over the same textured surface, performing MC localization on an offboard machine with 20 particles using the map created by SLAM. These poses are obtained at a rate of 14 Hz. We compare the pose data from offline SLAM with ground-truth obtained by a motion capture system. Table \ref{table:slamacc} gives the mean, standard deviation, maximum, and minimum error. Fig. \ref{fig:offboardSLAM} plots the pose estimate from offline SLAM compared to ground truth.   

\begin{table}[t]
  \centering
    \begin{tabular}{rcccc}
        \toprule
          & Mean & Std & Maximum & Minimum\\
        \midrule
        Offboard Hand-Held & 0.127 & 0.0677 & 0.265 & 0.012\\
        \bottomrule
    \end{tabular}
    \caption{Error between our SLAM implementation and ground truth from motion capture system (meters).}
    \label{table:slamacc}
    \vspace{-4mm}
\end{table}

\section{IMPACT IN EDUCATION}
% Since the first run of the PiDrone course during the Fall 2017 semester at Brown University, the PiDrone platform has broadly expanded its scope of influence. 

The first run of the PiDrone course by Brand et al. \cite{brand2018pidrone} demonstrated the potential of the platform and course. Simplified versions of the course have been used to introduce robotics to dozens of high school students. At the Providence Career and Technical Academy, the PiDrone has been used as part of the engineering curriculum for two years. The course has been taught at a Summer@Brown session to high school students from across the nation and world putting autonomous flight in the hands of students of diverse backgrounds. The course was also offered at a rural public school in Upstate New York, again proving the ability of the course to empower students of all backgrounds with their own autonomous robots. The PiDrone course for high school students will be taught again at Summer@Brown and at MassRobotics, a local robotics incubator in Boston, Massachusetts \cite{massrobotics}.

Increasing platform accessibility, the PiDrone platform has merged with Duckietown under the name of Duckiesky \cite{duckietown_text_only_trans}. Duckietown is a popular platform for learning autonomy on ground robots \cite{duckietown_text_only_trans}. This merge resulted in the creation of an online textbook which makes the platform, learning materials, and operation instructions readily accessible. These new online resources were used for the Fall 2018 course at Brown University. The course proved successful with 23/25 students covering the rigorous content and implementing their own algorithms for autonomy. The greatest challenge of the course is introducing a broad array of robotics concepts within a time frame that only allows students to scratch the surface. Despite this challenge, upon completion of the course, students are well-equipped with the tools needed to approach problems in autonomous robotics.

The future outlook of the platform as an educational tool includes expansion into additional universities as an undergraduate course, and the creation of online learning modules and additional projects for learning and building upon the drone's current capabilities.

%%%%%%%%%%%%%%%%%%%%%%%%%%%%%%%%%%%%%%%%%%%%%%%%%%%%%%%%%%%%%%%%%%%%%%%%%%%%%%%%

\section{FUTURE WORK}
The PiDrone is still under heavy development. We are still working on the drone's stability and safety, as well as increasing functionality.

We plan to continue expanding the PiDrone platform, particularly its autonomous capabilities.
Huang et al. have performed research ~\cite{huang19mrdrone} using the PiDrone with Mixed Reality, natural language commands, and high-level planning.
Adding high-level planning and support for Markov Decision Processes allows students to learn and work with reinforcement learning, which can build upon the work of Huang et al.
For hardware updates, we would like to replace the current IR sensor with a sensor that can more accurately estimate the drone's height at higher altitudes. Further exploration of adding a forward-facing camera to the drone will enable the implementation of SLAM in three dimensions.

The addition of a UKF, localization, and SLAM to the drone demonstrates that the drone can support higher level autonomy, and opens the door for implementing more advanced autonomous functionality. Continued work regarding the drone's vision capabilities will permit object tracking and motion planning tasks. Generating a proper dynamics model of the drone will allow for the implementation of advanced control algorithms that are better suited than a PID controller for performing aggressive maneuvers such as acrobatic flipping or perching. Based on the expanding capabilities of the platform, new instructional projects will be created for future iterations of the educational course, which will run again in Fall 2019 at Brown University.

The course will also be integrated into the edX platform ~\cite{edx} so that it can be taught both online and in residential settings. A crowdfunding campaign is planned to enable packaging of the drone parts into self-contained kits to distribute to individuals who desire to learn autonomous robotics using the PiDrone platform. With the modularized software architecture and existing capabilities of the drone, students, educators, and researchers alike can easily use and build upon the PiDrone platform to explore autonomy in aerial robotics.

%%%%%%%%%%%%%%%%%%%%%%%%%%%%%%%%%%%%%%%%%%%%%%%%%%%%%%%%%%%%%%%%%%%%%%%%%%%%%%%%

\section{CONCLUSION}
Current educational robots do not exhibit significant autonomous abilities. Advancing the PiDrone, we present a low-cost educational drone platform for an introductory robotics course that teaches advanced algorithms for state estimation in an accessible way. Although implemented on a Raspberry Pi in Python, the performance of the UKF, MC localization, and FastSLAM on the PiDrone makes the platform a compelling framework for introducing students of any background to robotics and high-level autonomy.

%%%%%%%%%%%%%%%%%%%%%%%%%%%%%%%%%%%%%%%%%%%%%%%%%%%%%%%%%%%%%%%%%%%%%%%%%%%%%%%%

%%%%%%%%%%%%%%%%%%%%%%%%%%%%%%%%%%%%%%%%%%%%%%%%%%%%%%%%%%%%%%%%%%%%%%%%%%%%%%%%

%%%%%%%%%%%%%%%%%%%%%%%%%%%%%%%%%%%%%%%%%%%%%%%%%%%%%%%%%%%%%%%%%%%%%%%%%%%%%%%%

\section*{ACKNOWLEDGMENT}
We would like to thank Amazon Robotics for their donations which help fund development and equipment for the course and platform.

This work is supported by the National Aeronautics and Space Administration under grant number NNX16AR61G.

We would like to thank Duckietown for their technical support.

We would also like to thank James Baccala, Jose Toribio, and Caesar Arita for their contributions to the drone hardware platform. Special thanks go to James for designing the propeller guards.

%%%%%%%%%%%%%%%%%%%%%%%%%%%%%%%%%%%%%%%%%%%%%%%%%%%%%%%%%%%%%%%%%%%%%%%%%%%%%%%%

%\bibliographystyle{plainnat}
\bibliographystyle{unsrtnat} % puts references in the order that they appear in the paper
\bibliography{IEEEabrv,references}

\end{document}